\documentclass[letterpaper]{article} 
\usepackage[draft]{aaai2026}  
\usepackage{times}  
\usepackage{helvet}  
\usepackage{courier}  
\usepackage[hyphens]{url}  
\usepackage{graphicx} 
\usepackage{natbib}  
\usepackage{caption} 
\usepackage{algorithm}
\usepackage{algorithmic}

\usepackage{amsmath}
\usepackage{amssymb}
\usepackage{color}

\usepackage{newfloat}
\usepackage{listings}
\DeclareCaptionStyle{ruled}{labelfont=normalfont,labelsep=colon,strut=off} 
\floatstyle{ruled}
\newfloat{listing}{tb}{lst}{}
\floatname{listing}{Listing}
\title{
A Computable Game-Theoretic Framework for Multi-Agent Theory of Mind
}
\author{
    Fengming Zhu,
    Yuxin Pan,
    Xiaomeng Zhu,
    Fangzhen Lin
}
\affiliations{
    The Hong Kong University of Science and Technology\\
    HongKong SAR, China\\
    \texttt{\{fzhuae, ypanav, xzhubr\}@connect.ust.hk\quad flin@cse.ust.hk}

}

\usepackage{bibentry}

\begin{document}

\maketitle

\begin{abstract}


Originating in psychology, \textit{Theory of Mind} (ToM) has attracted significant attention across multiple research communities, especially logic, economics, and robotics.
Most psychological work does not aim at formalizing those central concepts, namely \textit{goals}, \textit{intentions}, and \textit{beliefs}, to automate a ToM-based computational process, which, by contrast, has been extensively studied by logicians.
In this paper, we offer a different perspective by proposing a computational framework viewed through the lens of game theory.
On the one hand, the framework prescribes how to make boudedly rational decisions while maintaining a theory of mind about others (and recursively, each of the others holding a theory of mind about the rest); on the other hand, it employs statistical techniques and approximate solutions to retain computability of the inherent computational problem.

\end{abstract}

\section{Introduction}

The investigation of Theory of Mind (ToM) traces back to the seminal work of~\citeauthor{premack1978does}~\shortcite{premack1978does}. 
They showed that a chimpanzee can (1) understand a human experimenter's goals and intentions rather than merely imitate procedural actions, and (2) infer how one human thinks about another.
Despite of numerous subsequent psychological studies on animals and human infants~\cite{call1998distinguishing, call2004unwilling, call2008does}, we here draw the reader's attention to similar ideas that are witnessed and need to be further developed in the AI community, especially in the sub-area of \textit{autonomous agents}.
In both fields, the central concepts of interest are referred to as \textit{goals}, \textit{intentions} and \textit{beliefs}.


In the logical formalisms of automated planning~\cite{ghallab2004automated} for a single agent, goals are typically sets of desired fluents, and beliefs are represented as an agent's \textit{mental states} that can be repeatedly revised~\cite{alchourron1985logic,shoham1993agent,bolander2011epistemic}.
Intentions interact with goals and beliefs in complex ways~\cite{shoham2009logical}, and can be deliberately divided into the cases of \textit{intending an action} versus \textit{intending a state}~\cite{shoham2008multiagent}.
Those formalisms were later extended from single-agent to multi-agent settings, including concurrent planning like \textsc{ConGolog}~\cite{de2000congolog}, higher-order belief change by nested modal operators~\cite{wan2021general}, and general game playing~\cite{genesereth2013international,genesereth2014general}. 


In this paper, we will try to convince the readers that decision-theoretic and even game-theoretic models are also useful to serve as an appropriate computational framework for ToM, complementing the aforementioned logical formalisms.
The benefits mainly lie in two aspects: 1) agents therein can be utilitarian, i.e., will maximize their rewards by best responding to the environment as well as to the other agents; and 2) modern statistical and machine learning techniques can be effectively integrated to scale up to larger domains.
One may at last observe that these decision/game-theoretic models and logic formalisms exhibit a tight reciprocal relation.

Due to the page limit, we will focus on elaborating our proposed theoretic framework, leaving experiments to future work.
This framework is inspired by the cognitive hierarchy mentioned in~\cite{camerer2004cognitive}, and further utilizes statistical conjugates to enable efficient update of the hierarchy.
One can also view this work as extending~\cite{albrecht2015game} and~\cite{zhu2025single}
in a way that beliefs can be constructed automatically in a recursive manner.
We postpone some other related work to Appendix~\ref{app:related}.

%
%
%
%
%
%
%
%
%
%
%
%

\section{The Formal Framework}

The system where all the agents interact is modeled as a \textit{stochastic game}~\cite{shapley1953stochastic, littman1994markov}, given as a 5-tuple $\langle \mathcal{N},\mathcal{S}, \mathcal{A}, T, R\rangle$,
\begin{itemize}
	\item $\mathcal{N}$ is a finite set of $n$ agents.
	\item $\mathcal{S}$ is a finite set of environmental state.
	\item $\mathcal{A} = \mathcal{A}_1 \times \cdots \times \mathcal{A}_n$ is a set of joint actions, where $\mathcal{A}_i$ is the action set of agent $i$. We write $a_i$ as the action of agent $i$ and $a = (a_i, a_{-i})$ as the joint action.
	\item $T: \mathcal{S} \times \mathcal{A}_1 \times \cdots \mathcal{A}_n \mapsto \Delta(\mathcal{S})$ defines stochastic transitions across states.
	\item $R_i: \mathcal{S} \times \mathcal{A}_1 \times \cdots \mathcal{A}_n \mapsto \mathbb{R}$ denotes the immediate reward of agent $i$ in state $S$ under the effect of the joint-action $a$.
\end{itemize}
It is worth noting that the \textbf{\textit{goal}} concept in automated planning is now subsumed by (and must be compatible with) the reward structure.
For example, an agent should not be penalized for reaching a location if the underlying goal is to get there.
Each agent acts in its own interest by maximizing accumulated rewards $\mathbb{E}[\sum_t\gamma^tR_{i,t}]$, where $\gamma$ is a user-specified discount factor, and the reward notation is slightly reloaded to index timesteps.
The \textbf{\textit{intention}} of an agent is then defined as a \textit{policy} to achieve the utilitarian objective (i.e., the underlying \textit{goal}), which is also referred to as a \textit{strategy} under the game-theoretic context.
As~\citeauthor{zhu2025constant}~\shortcite{zhu2025constant} showed that formulations in stationary strategies can be perfectly carried over to strategies that takes in histories of finite lengths, we therefore focus on the stationary case in this paper, for the sake of notational simplicity.
A stationary policy of agent $i$ maps from environmental states to (potentially randomized) actions, i.e., $\pi_i: \mathcal{S} \mapsto \Delta(\mathcal{A}_i)$.
Nevertheless, an agent $i$ may not know another agent $j$'s reward structure, hence agent $j$'s exact intention remains unrevealed to agent $i$ and the others.
Consequently, the \textbf{\textit{belief}} that an agent holds is some sort of well-structured private information that can be used to infer the up-to-date \textit{intentions} of the others.

We formalize the \textbf{\textit{belief}} structure using the ${Poisson(\lambda)}$ cognitive hierarchy~\cite{camerer2004cognitive}, where we further assume the parameter $\lambda$ is drawn from a prior $Gamma(a,b)$ distribution to enable Bayesian inference.
The hierarchy is constructed level by level as follows,
\begin{itemize}
	\item Level-0 agents perform certain random strategies (or adhere to some simple, obvious rules).
	\item Recursively, level-$k$ agents best respond to those below level-$k$. We will discuss two implementations later.
\end{itemize}
Conceptually, a level-$\infty$ agent is a perfectly rational one.
The probability of an agent belonging to level-$k$ is assumed to be the following, with each agent modelled independently,
\[
f(k;\lambda) = e^{-\lambda} \frac{\lambda^{k}}{k!},\quad k=0,1,2,\cdots
\]
That is, the level $K$, as a discrete random variable, follows a $Poisson(\lambda)$ distribution.
Note that $\mathbb{E}[K] = \lambda$, capturing the average reasoning level of the population,
which is further assumed to be a continuous random variable $\Lambda$ drawn from a prior \textit{Gamma} distribution, i.e., $\Lambda \sim Gamma(a,b)$ with its expectation $\mathbb{E}(\Lambda) = \frac{a}{b}$.
After observing $m$ additional rounds of interactions yielding $(k_1, \cdots, k_m)$, where $k_r$ means the opponent has played a level-$k_r$ strategy at the $r$-th round, the posterior distribution can be updated to $Gamma(a + \sum_r k_r, b+m)$,
and the next best estimation of $\Lambda$ is
$\lambda' \gets E[\Lambda|(k_1, \cdots, k_m)] = \frac{a + \sum_r k_r}{b+m}$.
The detailed derivation of this Bayesian update of the Gamma-Poisson conjugacy is attached in Appendix~\ref{app:conj}.

We then elaborate how to construct the strategy in each level of the hierarchy. In the construction below, by $BR(\cdot)$ we mean to compute the best response, and by $\mathcal{M}(\cdot)$ we mean to induce a corresponding MDP~\cite{zhu2025single}. The computational details and pseudocode illustration are postponed to Appendix~\ref{app:br} and~\ref{app:pseudo}, respectively.

For each agent $j$, the 0-th level strategy is given by some ad-hoc rule, or follows a random distribution, e.g.,
\[
\forall S\in \mathcal{S},\ \pi_j|_0(a_j|S) \sim Unif(\mathcal{A}_j)
\]
As mentioned, there can be at least two ways of the bottom-up construction, both of which demonstrate some closure property under the $BR(\cdot)$ operator.
\begin{enumerate}
\item The agent in level-$(k+1)$ assumes all the others belong to level-$k$, and best responds to a singleton strategy, i.e.,
\[
\pi_j|_{k+1} \in BR(\pi_{-j}|_{k})
\]
which can be obtained by optimally solving the induced MDP $\mathcal{M}(\pi_{-j}|_k)$~\cite{zhu2025single}.
\textit{
In this case, if $\pi_j|_0$ is a stationary strategy, then $\pi_j|_{1} \in BR(\pi_{-j}|_{0})$ can be devised by optimally solving $\mathcal{M}(\pi_{-j}|_0)$, and therefore, is also stationary.
Inductively, up to any $k$, $\pi_j|_{k+1} \gets BR(\pi_{-j}|_{k})$ is also stationary.
}

\item The agent in level-$(k+1)$ assumes that the rest are  distributed over the levels no more than $k$ according to the normalized $Poisson$ distribution that is imposed globally, and therefore, best responds to a mixed strategy, i.e.,
\[
\pi_j|_{k+1} \in 
BR(\pi_{-j}^{mixed}|_{k})
\vspace{-0.6mm}
\]
where $\pi_{-j}^{mixed}|_{k} \gets \pi_{-j}|_\iota$ w.p. $g_\iota = f_\lambda(\iota)/\sum_{\iota'=0}^{k} f_\lambda(\iota')$,
which coincides with the conditional probability given that the reasoning level is at most $k$.

In principle, to compute $BR(\pi_{-j}^{mixed}|_{k})$~\cite{zhu2025single}, one has to solve an underlying POMDP~\cite{sondik1978optimal,kaelbling1998planning}, which will incur potential undecidability~\cite{madani2003undecidability}.
To retain computability, we leveraged the QMDP approximation~\cite{littman1995learning}, given as,
\[
\pi_j|_{k+1}(S) \in \arg\max_{a_j \in \mathcal{A}_j} \sum_{\iota=0}^k g_\iota \cdot Q^*_{\mathcal{M}(\pi_{-j}|_\iota)}(S, a_j)
\]
where $Q^*_{\mathcal{M}(\pi_{-j}|_\iota)}$ is the Q-function obtained by optimally solving the MDP $\mathcal{M}(\pi_{-j}|_\iota)$.
\textit{
One can see that if $\pi_j|_0$ is a stationary strategy, then up to any $k$,
	$\pi_j|_{k+1} \in BR(\pi_{-j}^{mixed}|_{k})$ is also a stationary policy.
}
\end{enumerate}
Note that the complexities of the above two implementations are different.
Suppose agent $j$ believes it is in level $K_j$.
For the first implementation, it needs to solve $\Theta(K_j)$ MDPs solely for the initial step.
Only the update of belief distribution is required, as the support strategies at each level of the hierarchy will remain the same.
For the second implementation, it needs to solve $\Theta(K_j)$ MDPs every time it updates the belief, since changes in beliefs (more specifically,  $g_\iota$'s) will incur changes in best responses.

To some extent, this framework can be viewed as an instantiation of I-POMDP~\cite{gmytrasiewicz2005framework}, but circumvents the latter's computability issues.
It is also important to note that, the belief update may be non-monotonic and non-converging, in terms of approaching a specific distribution with decreasing entropies, as sometimes the belief may be easily restored when the it does not accurately reflect the truth~\cite{albrecht2019convergence}.

\section{Concluding Remarks}

In this paper, we propose a practical framework that embeds and formalizes those central concepts in ToM, while the inherent computability is still remained.
With the help of this framework,
the behavior of a population of agents, robots, or humans can be modeled and prescribed.
Some experimental results conducted on certain human-robot cohabiting systems will be reported soon.

\bibliography{aaai2026}

\begin{thebibliography}{32}
\providecommand{\natexlab}[1]{#1}

\bibitem[{Albrecht and Ramamoorthy(2015)}]{albrecht2015game}
Albrecht, S.~V.; and Ramamoorthy, S. 2015.
\newblock A game-theoretic model and best-response learning method for ad hoc
  coordination in multiagent systems.
\newblock \emph{arXiv preprint arXiv:1506.01170}.

\bibitem[{Albrecht and Ramamoorthy(2019)}]{albrecht2019convergence}
Albrecht, S.~V.; and Ramamoorthy, S. 2019.
\newblock On convergence and optimality of best-response learning with policy
  types in multiagent systems.
\newblock \emph{arXiv preprint arXiv:1907.06995}.

\bibitem[{Alchourr{\'o}n, G{\"a}rdenfors, and
  Makinson(1985)}]{alchourron1985logic}
Alchourr{\'o}n, C.~E.; G{\"a}rdenfors, P.; and Makinson, D. 1985.
\newblock On the logic of theory change: Partial meet contraction and revision
  functions.
\newblock \emph{The journal of symbolic logic}, 50(2): 510--530.

\bibitem[{Bolander and Andersen(2011)}]{bolander2011epistemic}
Bolander, T.; and Andersen, M.~B. 2011.
\newblock Epistemic planning for single-and multi-agent systems.
\newblock \emph{Journal of Applied Non-Classical Logics}, 21(1): 9--34.

\bibitem[{Boutilier(1996)}]{boutilier1996planning}
Boutilier, C. 1996.
\newblock Planning, learning and coordination in multiagent decision processes.
\newblock In \emph{TARK}, volume~96, 195--210.

\bibitem[{Call et~al.(2004)Call, Hare, Carpenter, and
  Tomasello}]{call2004unwilling}
Call, J.; Hare, B.; Carpenter, M.; and Tomasello, M. 2004.
\newblock `Unwilling' versus `unable': chimpanzees' understanding of human
  intentional action.
\newblock \emph{Developmental science}, 7(4): 488--498.

\bibitem[{Call and Tomasello(1998)}]{call1998distinguishing}
Call, J.; and Tomasello, M. 1998.
\newblock Distinguishing intentional from accidental actions in orangutans
  (Pongo pygmaeus), chimpanzees (Pan troglodytes) and human children (Homo
  sapiens).
\newblock \emph{Journal of comparative psychology}, 112(2): 192.

\bibitem[{Call and Tomasello(2008)}]{call2008does}
Call, J.; and Tomasello, M. 2008.
\newblock Does the chimpanzee have a theory of mind? 30 years later.
\newblock \emph{Trends in cognitive sciences}, 12(5): 187--192.

\bibitem[{Camerer, Ho, and Chong(2004)}]{camerer2004cognitive}
Camerer, C.~F.; Ho, T.-H.; and Chong, J.-K. 2004.
\newblock A cognitive hierarchy model of games.
\newblock \emph{The quarterly journal of economics}, 119(3): 861--898.

\bibitem[{De~Giacomo, Lesp{\'e}rance, and Levesque(2000)}]{de2000congolog}
De~Giacomo, G.; Lesp{\'e}rance, Y.; and Levesque, H.~J. 2000.
\newblock ConGolog, a concurrent programming language based on the situation
  calculus.
\newblock \emph{Artificial Intelligence}, 121(1-2): 109--169.

\bibitem[{Genesereth and Bj{\"o}rnsson(2013)}]{genesereth2013international}
Genesereth, M.; and Bj{\"o}rnsson, Y. 2013.
\newblock The international general game playing competition.
\newblock \emph{AI Magazine}, 34(2): 107--107.

\bibitem[{Genesereth and Thielscher(2014)}]{genesereth2014general}
Genesereth, M.; and Thielscher, M. 2014.
\newblock \emph{General game playing}.
\newblock Morgan \& Claypool Publishers.

\bibitem[{Ghallab, Nau, and Traverso(2004)}]{ghallab2004automated}
Ghallab, M.; Nau, D.; and Traverso, P. 2004.
\newblock \emph{Automated Planning: theory and practice}.
\newblock Elsevier.

\bibitem[{Gmytrasiewicz and Doshi(2005)}]{gmytrasiewicz2005framework}
Gmytrasiewicz, P.~J.; and Doshi, P. 2005.
\newblock A framework for sequential planning in multi-agent settings.
\newblock \emph{Journal of Artificial Intelligence Research}, 24: 49--79.

\bibitem[{Jin et~al.(2024)Jin, Wu, Cao, Xiang, Kuo, Hu, Ullman, Torralba,
  Tenenbaum, and Shu}]{jin2024mmtom}
Jin, C.; Wu, Y.; Cao, J.; Xiang, J.; Kuo, Y.-L.; Hu, Z.; Ullman, T.; Torralba,
  A.; Tenenbaum, J.; and Shu, T. 2024.
\newblock MMToM-QA: Multimodal Theory of Mind Question Answering.
\newblock In \emph{Proceedings of the 62nd Annual Meeting of the Association
  for Computational Linguistics (Volume 1: Long Papers)}, 16077--16102.

\bibitem[{Kaelbling, Littman, and Cassandra(1998)}]{kaelbling1998planning}
Kaelbling, L.~P.; Littman, M.~L.; and Cassandra, A.~R. 1998.
\newblock Planning and acting in partially observable stochastic domains.
\newblock \emph{Artificial intelligence}, 101(1-2): 99--134.

\bibitem[{Littman(1994)}]{littman1994markov}
Littman, M.~L. 1994.
\newblock Markov games as a framework for multi-agent reinforcement learning.
\newblock In \emph{Machine learning proceedings 1994}, 157--163. Elsevier.

\bibitem[{Littman, Cassandra, and Kaelbling(1995)}]{littman1995learning}
Littman, M.~L.; Cassandra, A.~R.; and Kaelbling, L.~P. 1995.
\newblock Learning policies for partially observable environments: Scaling up.
\newblock In \emph{Machine Learning Proceedings 1995}, 362--370. Elsevier.

\bibitem[{Madani, Hanks, and Condon(2003)}]{madani2003undecidability}
Madani, O.; Hanks, S.; and Condon, A. 2003.
\newblock On the undecidability of probabilistic planning and related
  stochastic optimization problems.
\newblock \emph{Artificial Intelligence}, 147(1-2): 5--34.

\bibitem[{Premack and Woodruff(1978)}]{premack1978does}
Premack, D.; and Woodruff, G. 1978.
\newblock Does the chimpanzee have a theory of mind?
\newblock \emph{Behavioral and brain sciences}, 1(4): 515--526.

\bibitem[{Rabinowitz et~al.(2018)Rabinowitz, Perbet, Song, Zhang, Eslami, and
  Botvinick}]{rabinowitz2018machine}
Rabinowitz, N.; Perbet, F.; Song, F.; Zhang, C.; Eslami, S.~A.; and Botvinick,
  M. 2018.
\newblock Machine theory of mind.
\newblock In \emph{International conference on machine learning}, 4218--4227.
  PMLR.

\bibitem[{Shapley(1953)}]{shapley1953stochastic}
Shapley, L.~S. 1953.
\newblock Stochastic games.
\newblock \emph{Proceedings of the national academy of sciences}, 39(10):
  1095--1100.

\bibitem[{Shi et~al.(2025)Shi, Ye, Fang, Jin, Isik, Kuo, and Shu}]{shi2025muma}
Shi, H.; Ye, S.; Fang, X.; Jin, C.; Isik, L.; Kuo, Y.-L.; and Shu, T. 2025.
\newblock Muma-tom: Multi-modal multi-agent theory of mind.
\newblock In \emph{Proceedings of the AAAI Conference on Artificial
  Intelligence}, volume~39, 1510--1519.

\bibitem[{Shoham(1993)}]{shoham1993agent}
Shoham, Y. 1993.
\newblock Agent-oriented programming.
\newblock \emph{Artificial intelligence}, 60(1): 51--92.

\bibitem[{Shoham(2009)}]{shoham2009logical}
Shoham, Y. 2009.
\newblock Logical theories of intention and the database perspective.
\newblock \emph{Journal of Philosophical Logic}, 38(6): 633--647.

\bibitem[{Shoham and Leyton-Brown(2008)}]{shoham2008multiagent}
Shoham, Y.; and Leyton-Brown, K. 2008.
\newblock \emph{Multiagent systems: Algorithmic, game-theoretic, and logical
  foundations}.
\newblock Cambridge University Press.

\bibitem[{Sondik(1978)}]{sondik1978optimal}
Sondik, E.~J. 1978.
\newblock The optimal control of partially observable Markov processes over the
  infinite horizon: Discounted costs.
\newblock \emph{Operations research}, 26(2): 282--304.

\bibitem[{Wan, Fang, and Liu(2021)}]{wan2021general}
Wan, H.; Fang, B.; and Liu, Y. 2021.
\newblock A general multi-agent epistemic planner based on higher-order belief
  change.
\newblock \emph{Artificial Intelligence}, 301: 103562.

\bibitem[{Wen, Yang, and Wang(2021)}]{wen2021modelling}
Wen, Y.; Yang, Y.; and Wang, J. 2021.
\newblock Modelling bounded rationality in multi-agent interactions by
  generalized recursive reasoning.
\newblock In \emph{Proceedings of the Twenty-Ninth International Conference on
  International Joint Conferences on Artificial Intelligence}, 414--421.

\bibitem[{Zhang et~al.(2025)Zhang, Jin, Jia, and Shu}]{zhang2025autotom}
Zhang, Z.; Jin, C.; Jia, M.~Y.; and Shu, T. 2025.
\newblock AutoToM: Automated Bayesian Inverse Planning and Model Discovery for
  Open-ended Theory of Mind.
\newblock In \emph{ICLR 2025 Workshop on Foundation Models in the Wild}.

\bibitem[{Zhu and Lin(2025{\natexlab{a}})}]{zhu2025constant}
Zhu, F.; and Lin, F. 2025{\natexlab{a}}.
\newblock Constant-Memory Strategies in Stochastic Games: Best Responses and
  Equilibria.
\newblock \emph{arXiv preprint arXiv:2505.07008}.

\bibitem[{Zhu and Lin(2025{\natexlab{b}})}]{zhu2025single}
Zhu, F.; and Lin, F. 2025{\natexlab{b}}.
\newblock Single-Agent Planning in a Multi-Agent System: A Unified Framework
  for Type-Based Planners.
\newblock In \emph{Proceedings of the 24th International Conference on
  Autonomous Agents and Multiagent Systems}, 2382--2391.

\end{thebibliography}

\clearpage
\appendix

\section*{Appendix}

\section{More Related Work on ToM}
\label{app:related}

Here we include additional related work on ToM with statistical/deep learning involved.
\citeauthor{boutilier1996planning}~\shortcite{boutilier1996planning} investigated the use of Dirichlet-Categorical conjugacy in opponent modelling, but it was only applied to tiny-scale matrix games. 
The machinery version of ToM was first proposed in~\cite{rabinowitz2018machine} leveraging POMDPs as the underlying model, along with proof-of-concept experiments in gird-worlds. However, the authors only covered modelling beliefs in a single-agent domain, without incorporating belief changes into planning in multi-agent domains.
The GR2 framework also employs the level-k theory of cognitive hierarchy~\cite{wen2021modelling}, which embeds a static $Poisson$ hierarchy within the RL training and performs no belief update during execution; and more subtly, it only considers \textit{action-wise} best responses (in any given state). In contrast, our approach updates the belief structure periodically and takes \textit{strategy-wise} best responses into account.
Also, as LLMs are increasingly prevalent, several benchmarks have emerged for evaluating the ToM abilities of LLMs, including sing-agent multimodal domains~\cite{jin2024mmtom}, multi-agent multimodal domains~\cite{shi2025muma}, and open-ended tasks~\cite{zhang2025autotom}. Yet, these work still does not present well-formalized terms for multi-agent ToM.

\section{Gamma-Poisson Conjugacy}
\label{app:conj}

The average reasoning level $\Lambda$ is assumed as a continuous random variable following a prior $Gamma$ distribution, i.e., $\Lambda \sim Gamma(a,b)$, given as (with the normalization term omitted in the second line),
\[
\begin{split}
f_\Lambda(\lambda)
&=\frac{1}{\Gamma(a)}(b\lambda)^ae^{-b\lambda}\frac{1}{\lambda}
\propto \lambda^{a-1} e^{-b\lambda}
,\quad \lambda > 0 \\
\end{split}
\]
The expectation is $\mathbb{E}[\Lambda] = \frac{a}{b}$.

Conditioning on a realized $\lambda \sim \Lambda$, we take the likelihood (sampling) distribution as a $Poisson$ distribution. Recall that $K$ denotes the random variable corresponding to the level of rational reasoning, with $K|\lambda \sim Possion(\lambda)$,
\[
\begin{split}
f_{K|\Lambda}(k|\lambda)
&= e^{-\lambda} \frac{\lambda^{k}}{k!}
\propto e^{-\lambda}\lambda^k ,\quad k=0,1,2,\cdots
\end{split}
\]
After observing $m$ further rounds of interaction and having $\vec{k} = (k_1, \cdots, k_m)$ (where $k_r$ means the opponent played a strategy corresponding to the $k_r$-th level of rationality at round $r$), the posterior distribution can be updated as,
\[
\begin{split}
f_{\Lambda|K}(\lambda|\vec{k})
& = \frac{f_{K|\Lambda}(\vec{k}|\lambda)f_\Lambda(\lambda)}{f_K(\vec{k})} \\ 
& = \frac{\prod_{r=1}^m \frac{e^{\lambda}\lambda^{k_r}}{k_r!} \cdot \frac{b^a}{\Gamma(a)}\lambda^{a-1}e^{-b\lambda}}{\int_0^\infty f_{K|\Lambda}(\vec{k}|\lambda)f_\Lambda(\lambda)\mathbf{d}\lambda} \\
& \propto e^{-m\lambda} \lambda^{k_1 + \cdots + k_m} \lambda^{a-1} e^{-b\lambda} \\
& = \lambda^{k_1 + \cdots + k_m + a - 1} e^{-(m+b)\lambda} \\
\end{split}
\]
which coincides with the form of the density function of $Gamma(a+\sum_r k_r, b+m)$,
and therefore, the next best prediction for $\Lambda$ is
$\lambda' \gets \mathbb{E}[\Lambda|K] = \frac{a+\sum_r k_r}{b+m}$.

\section{Best Response Oracles}
\label{app:br}
\citeauthor{zhu2025constant}~\shortcite{zhu2025constant} have provided a rigorous proof, so here we just echo some intuitive justifications.
For best responding to a single strategy (profile), 
if it is provided that the opponents are playing a stationary strategy profile $\pi_{-j}$, then from agent $j$'s perspective, she is faced with an induced MDP $\mathcal{M}(\pi_{-j}) = \langle \mathcal{S}, \mathcal{A}_j, T^{\pi_{-j}}, R^{\pi_{-j}}, \gamma \rangle$, defined as,
	\begin{itemize}
		\item $\mathcal{S}, \mathcal{A}_j$ and $\gamma$ is carried over from the previous setup,
		\item $T^{\pi_{-j}}(S'| S, a_j) \triangleq \sum_{a_{-j} \in \mathcal{A}_{-j}}T(S'|S,a)\pi_{-j}(a_{-j}|S)$,
		\item $R^{\pi_{-j}}(S, a_j) \triangleq \sum_{a_{-j} \in \mathcal{A}_{-j}}R_i(S,a)\pi_{-j}(a_{-j}|S)$,
	\end{itemize}
Hence, the best response $BR(\pi_{-j})$ is to optimally solve the induced MDP $\mathcal{M}(\pi_{-j})$, where there must exist an optimal stationary policy serving as a stationary (and pure) strategy best response .

\section{Algorithmic Procedures}
\label{app:pseudo}

\begin{figure}[!ht]
	\centering
	\includegraphics[width=49mm]{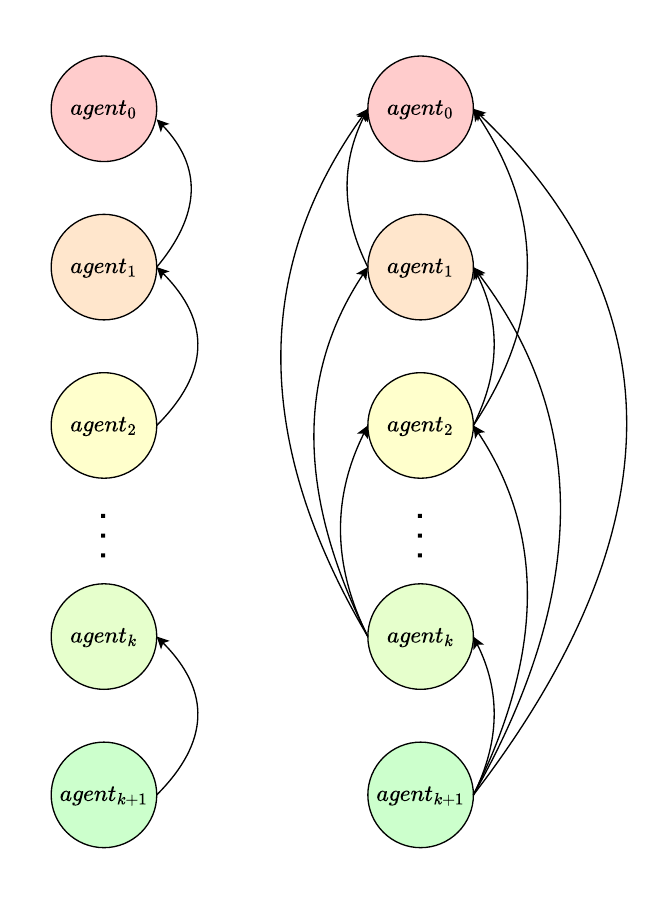}
	\caption{An illustration of the hierarchical ToM framework for the two implementations, respectively. $agent_\iota$ means the agent (or the population of agents) residing in level-$\iota$.}
	\label{app:fig:tom}
\end{figure}

For a modelling agent $i$, its job is twofold: (1) computing a belief structure of the other agents being modelled, denoted as $-i$; (2) treating the computed belief structure as a mixed strategy hold by the others and computing a best response. The detailed procedure is as follows,  
\begin{enumerate}
	\item Begin with an assumed prior $Gamma(a,b)$.
	\item Repeat the following steps,
	\begin{enumerate}
		\item Estimate a $\lambda$ for the $Poisson(\lambda)$ belief distribution.
		\item Adopt one of the two aforementioned implementations, also illustrated in Figure~\ref{app:fig:tom}, to compute the support strategies of the belief structure. \textit{For the first implementation, this is only needed in the first iteration.}
		\item Compute its own best response strategy against this belief structure.
		\item Observe the others' played strategies, and update the parameters of the prior, namely $(a, b)$, accordingly.
	\end{enumerate}
\end{enumerate}

\end{document}